%% file: main.tex
\documentclass[runningheads]{llncs}

\usepackage{amsmath, amssymb, amsfonts}
\usepackage{url}
\usepackage{xcolor}
\usepackage{hyperref}
\usepackage{graphicx}
\usepackage{verbatim}
\usepackage{latexsym}
\usepackage{setspace}
\renewcommand{\vec}[1]{\mathbf{#1}}
\newtheorem{prop}{Proposition}

\DeclareMathOperator{\rank}{rank}
\DeclareMathOperator{\trace}{Tr}
\DeclareMathOperator*{\argmax}{arg\,max}

\graphicspath{{figures/}}

\begin{document}
\title{Jacobian Regularization for Mitigating Universal Adversarial Perturbations\thanks{Kenneth T. Co is supported in part by the DataSpartan research grant DSRD201801.
The final authenticated version is available online at \url{https://doi.org/10.1007/978-3-030-86380-7_17}.}}
\author{Kenneth T. Co\inst{1,2}\orcidID{0000-0003-2766-7326} \and
David Martinez Rego\inst{2}\orcidID{0000-0003-1809-1169} \and
Emil C. Lupu\inst{1}\orcidID{0000-0002-2844-3917}}
\authorrunning{K. Co et al.}
\institute{Imperial College London, London SW7 2AZ, United Kingdom
\email{\{k.co,e.c.lupu\}@imperial.ac.uk}\\
\and DataSpartan, London EC2Y 9ST, United Kingdom\\
\email{david@dataspartan.com}}
\maketitle       
\begin{abstract}
Universal Adversarial Perturbations (UAPs) are input perturbations that can fool a neural network on large sets of data. They are a class of attacks that represents a significant threat as they facilitate realistic, practical, and low-cost attacks on neural networks. In this work, we derive upper bounds for the effectiveness of UAPs based on norms of data-dependent Jacobians. We empirically verify that Jacobian regularization greatly increases model robustness to UAPs by up to four times whilst maintaining clean performance. Our theoretical analysis also allows us to formulate a metric for the strength of shared adversarial perturbations between pairs of inputs. We apply this metric to benchmark datasets and show that it is highly correlated with the actual observed robustness. This suggests that realistic and practical universal attacks can be reliably mitigated without sacrificing clean accuracy, which shows promise for the robustness of machine learning systems.

\end{abstract}

\input{sections/1-introduction}
\input{sections/2-background}
\input{sections/3-formulation}
\input{sections/4-experiments}
\input{sections/5-conclusion}

\bibliographystyle{splncs04}
\bibliography{main}

\end{document}

%% file: sections/1-introduction.tex
\section{Introduction}
Neural networks have been the algorithm of choice for many applications such as image classification \cite{krizhevsky2012imagenet}, real-time object detection \cite{redmon2016you}, and speech recognition \cite{hinton2012deep}. Although they appear to be robust to noise, their accuracy can rapidly deteriorate in the face of adversarial examples -- inputs that appear similar to genuine data, but have been maliciously designed to fool the model \cite{biggio2013evasion,szegedy2014intriguing}. Thus, it is important to ensure that neural networks are robust to such attacks, especially in safety-critical applications, as this can greatly undermine the performance and trust in these models.

A concerning subset of attacks on neural networks come in the form of Universal Adversarial Perturbations (UAPs), where a single adversarial perturbation can cause a model to misclassify a large set of inputs \cite{moosavi2017universal}. These present a systemic risk, as many practical and physically realizable adversarial attacks are based on UAPs. These attacks can take the form of adversarial patches for image classification \cite{brown2017adversarial}, person recognition \cite{thys2019fooling}, camera-based \cite{eykholt2018physical,eykholt2018robust} and LiDAR-based object detection \cite{cao2019adversarial,hau2021object,hau2020ghostbuster,tu2020physically}. In the digital domain, UAPs have been shown to facilitate realistic attacks on perceptual ad-blockers for web pages \cite{tramer2019adversarial} and machine learning-based malware detectors \cite{labaca2021universal}. Furthermore, an attacker can utilize UAPs to perform query-efficient black-box attacks on neural networks \cite{co2019procedural,co2019sensitivity}.

In the literature, existing defenses to adversarial attacks focus primarily on input-specific (``per-input'') attacks--where adversarial perturbations need to be crafted \emph{for each single input}. In contrast to universal attacks, input-specific attacks fool the model on only \emph{one input}. However, the practicality of input-specific attacks suffers in realistic settings, as the perturbations need to be constantly modified to match the current input. In contrast, defences against UAPs have not been thoroughly investigated, even if they are potentially more dangerous and should intuitively be easier to defend against because the same perturbation needs to be shared across many inputs. These are the main focus of this paper.

 A number of studies have investigated the use of Jacobian regularization to improve the stability of model predictions to small changes to the input, but up to this point, studies have only considered input-specific perturbations \cite{hoffman2019robust,jakubovitz2018improving,novak2018sensitivity,roth2020adversarial,sokolic2017robust,varga2017gradient}. In this work, we expand the theoretical formulation of Jacobian regularization to UAPs and derive upper bounds on the effectiveness of UAPs based on the properties of Jacobian matrices for individual inputs. Our work shows that for inputs to strongly share adversarial perturbations, their Jacobians need to share singular vectors.

We empirically verify our theoretical findings by applying Jacobian regularization to neural networks trained on popular benchmark datasets: MNIST \cite{lecun1998gradient}, Fashion-MNIST \cite{xiao2017fashion} and then evaluating their robustness to various UAPs. Our results show that even a small amount of Jacobian regularization drastically improves model robustness against many universal attacks with negligible downsides to clean performance. To summarize, we make the following contributions:

\begin{itemize}
    \item We extend theoretical formulations for universal adversarial perturbations and are the first to show that the effectiveness of UAPs is bounded above by the norms of data-dependent Jacobians.
    
    \item We empirically verify our theoretical results and show that even a minimal amount of Jacobian regularization reduces effectiveness of UAPs by up to 4-times, whilst leaving clean accuracy relatively unaffected.
    
    \item We propose the use of cosine similarity for Jacobians of inputs to measure the strength of shared adversarial perturbations between distinct inputs. Our empirical evaluations on benchmark datasets demonstrate that this similarity measure is an effective proxy for measuring robustness to UAPs.
\end{itemize}

The rest of this paper is organized as follows. Section~II introduces adversarial examples, universal adversarial perturbations, and Jacobian regularization. Section~III formulates Jacobian regularization for UAPs and derives our key propositions. Section~IV evaluates the robustness of models trained with Jacobian regularization to various UAP attack. Finally, Section~V discusses implications of our results and summarizes our findings.

%% file: sections/2-background.tex
\section{Background}

\subsection{Universal Adversarial Perturbations}
Let $f: \mathcal{X} \subset  \mathbb{R}^n \to \mathbb{R}^d$ denote the logits of a piece-wise linear classifier which takes as input $\vec{x} \in \mathcal{X}$. The output label assigned by this classifier is defined by $F(\vec{x}) = \argmax(f(\vec{x}))$. Let $\tau(\vec{x)}$ denote the true class label of an input $\vec{x)}$.

An \emph{adversarial example} $\vec{x}'$ is an input that satisfies $F(\vec{x}') \neq \tau(\vec{x})$, despite $\vec{x}'$ being close to $\vec{x}$ according to some distance metric (implicitly, $\tau(\vec{x}) = \tau(\vec{x}')$). The difference $\delta = \vec{x}' - \vec{x}$ is referred to as an adversarial perturbation and its norm is often constrained to $\Vert \delta \Vert_p < \varepsilon$, for some $\ell_p$-norm and small $\varepsilon > 0$ \cite{szegedy2014intriguing}.

\textbf{Universal Adversarial Perturbations (UAP)} can come in targeted or untargeted forms depending on the attacker's objective. An untargeted UAP is an adversarial perturbation $\delta \in \mathbb{R}^n$ that satisfies $F(\vec{x} + \delta) \neq \tau(\vec{x})$ for sufficiently many $\vec{x} \in \mathcal{X}$ and with $\Vert \delta \Vert_p < \varepsilon$ \cite{moosavi2017universal}. Untargeted UAPs are generated by maximizing the loss $\sum_i \mathcal{L}(\vec{x}_i + \delta)$ with an iterative stochastic gradient descent algorithm \cite{co2019universal,shafahi2018universal,mummadi2019defending,tramer2019adversarial}. Here, $\mathcal{L}$ is the model's training loss, $\{\vec{x}_i\}$ are batches of inputs, and $\delta$ are small perturbations that satisfy $\Vert \delta \Vert_p < \varepsilon$. Updates to $\delta$ are done in mini-batches in the direction of $-\sum_i \nabla \mathcal{L}(\vec{x}_i + \delta)$. Targeted UAPs for a class $c$ are adversarial perturbations $\delta$ that satisfy $F(\vec{x} + \delta) = c$ for sufficiently many $\vec{x} \in \mathcal{X}$ and with $\Vert \delta \Vert_p < \varepsilon$. To generate this type of attack, we use the same stochastic gradient descent as in the untargeted case, but modify the loss to be minimized when all resulting inputs $\vec{x}_i + \delta$ are classified as $c$.

\subsection{Jacobian Regularization}
Given that $f(\vec{x})$ is the logit output of the classifier for input $\vec{x}$, we write $\vec{J}_{f}(\vec{x})$ to denote the input-output Jacobian of $f$ at $\vec{x}$. We can linearise $f$ within a neighbourhood around $\vec{x}$ as follows using the Taylor series expansion:
\begin{equation}
f(\vec{x} + \delta) = f(\vec{x}) + \vec{J}_{f}(\vec{x}) \delta + O(\delta^2)
\end{equation}
For a sufficiently small neighbourhood $\Vert \delta \Vert_p \leq \varepsilon$ with $\varepsilon > 0$, the higher order terms of $\delta$ can be neglected and the stability of the prediction is determined by the Jacobian.
\begin{equation}\label{eq:approx}
f(\vec{x} + \delta) \simeq f(\vec{x}) + \vec{J}_{f}(\vec{x}) \delta
\end{equation}
and equivalently, for any $q$-norm, we have:
\begin{equation}\label{eq:norm}
\Vert f(\vec{x} + \delta) -  f(\vec{x}) \Vert_q \approx \Vert \vec{J}_{f}(\vec{x}) \delta \Vert_q
\end{equation}
For a small $\varepsilon$, we want the $\delta$ that maximizes the right hand side of Eq.~\ref{eq:norm} in order to sufficiently change the original output and fool the model. With constraint $\Vert \delta \Vert_p \leq \varepsilon$, this is equivalent to finding the $(p, q)$ singular vector for $\vec{J}_{f}(\vec{x})$ \cite{khrulkov2018art}.

To improve the stability of model outputs to small perturbations $\delta$, existing works have proposed regularizing the Frobenius norm \cite{hoffman2019robust,jakubovitz2018improving,novak2018sensitivity} or the Spectral norm \cite{roth2020adversarial,sokolic2017robust,varga2017gradient} of this data-dependent Jacobian $\vec{J}_{f}(\vec{x})$ for each input. Additionally, \cite{roth2020adversarial} show that the input-specific adversarial perturbations align with the dominant singular vectors of these Jacobian matrices.

Although \cite{khrulkov2018art} considered Jacobians in the context of UAPs, they only focused on the computation of $\delta$ as an attack and did not perform any theoretical or empirical analysis for mitigating the effects of UAPs. Prior studies that explore Jacobian regularization focused solely on improving robustness to single-input perturbations and did not explain nor consider the effectiveness of Jacobian regularization for UAPs. Thus, we extend these formulations \cite{khrulkov2018art,roth2020adversarial} to have a more concrete theoretical understanding for how Jacobian regularization mitigates UAPs.

%% file: sections/3-formulation.tex
\section{Jacobians for Universal Adversarial Perturbations}
When computing a universal adversarial perturbation $\delta$ that uniformly generalizes across multiple inputs $\{\vec{x}_i\}_{i = 1}^N$, one would optimize:
\begin{equation}
\label{eq:pre-stacked}
\max_{\delta: \Vert \delta \Vert_p = 1} \sum_{i = 1}^N \Vert \vec{J}_{f}(\vec{x}_i) \delta \Vert_q
\end{equation}
This extends the intuition from Eq.~\ref{eq:norm} to many inputs, and due to the homogeneity of the norm, it is sufficient to solve this for $\Vert \delta \Vert_p = 1$ \cite{khrulkov2018art}. The solution to $\delta$ for Eq.~\ref{eq:pre-stacked} is equivalent to finding the $(p, q)$ singular vector for the \textbf{stacked Jacobian} matrix $\overline{\vec{J}}_N$, the matrix formed by vertically stacking the Jacobians of the first $N$ inputs.
\begin{equation}
\label{eq:operator-uap}
\max_{\delta: \Vert \delta \Vert_p = 1} \Vert \overline{\vec{J}}_N \delta \Vert_q
\quad \text{ where } \quad
\overline{\vec{J}}_N =
    \begin{bmatrix}
    \vec{J}_{f}(\vec{x}_1)  \\
    \vec{J}_{f}(\vec{x}_2)  \\
    \vdots \\
    \vec{J}_{f}(\vec{x}_N)  \\
    \end{bmatrix}
\end{equation}

\subsection{Upper Bounds for the Stacked Jacobian}
To obtain an upper bound for the $(p, q)$-operator norm shown in Eq.~\ref{eq:operator-uap}, note that it is bounded above by its Frobenius norm denoted by $\Vert \overline{\vec{J}}_N \Vert_F$:
\begin{equation}
\Vert \overline{\vec{J}}_N \delta \Vert_q
\leq \Vert \overline{\vec{J}}_N \Vert_F \Vert\delta \Vert
\end{equation}
Thus, mitigating the effectiveness of a UAP across multiple inputs can be achieved by limiting the Frobenius norm of the stacked Jacobian $\Vert \overline{\vec{J}}_N \Vert_F$.

Before proceeding, let us define  the inner product induced by the Frobenius norm for two real matrices. Given $\vec{A}, \vec{B} \in \mathbb{R}^{m \times n}$, let the inner product in $\mathbb{R}^{m \times n}$ be defined as:
    \begin{equation}
    \langle \vec{A}, \vec{B} \rangle = \trace(\vec{A}'\vec{B}) = \sum_{i = 1}^m \sum_{j = 1}^n a_{ij} b_{ij}
    \end{equation}
where $\vec{A}'$ denotes the transpose of $\vec{A}$, the lowercase letters $a_{ij}$ are the entries of the matrix $\vec{A}$, and $\trace(\cdot)$ is the trace. This inner product is associated with the Frobenius norm $\Vert \cdot \Vert_F$. Now we introduce the following proposition.

\begin{prop}\label{prop:frob}
For matrices $\vec{A}, \vec{B} \in \mathbb{R}^{m \times n}$, we have:
    \begin{equation}
    \label{eq:alignment}
    \langle \vec{A}, \vec{B} \rangle \leq \Vert \vec{A} \Vert_F \Vert \vec{B} \Vert_F
    \end{equation}
with equality if and only if $\vec{A}$ and $\vec{B}$ share singular directions and their singular values satisfy $\sigma_i(\vec{A}) = s \cdot \sigma_i(\vec{B})$ for all $i$ for a constant scalar $s > 0$, where $\sigma_i(\cdot)$ is the singular value that corresponds to the $i$-th largest singular value.
\end{prop}

\begin{proof}
    Consider the singular value decomposition of $\vec{A} = \vec{U}_A \vec{\Sigma}_A \vec{V}_A'$ and $\vec{B} = \vec{U}_B \vec{\Sigma}_B \vec{V}_B'$, where $\vec{U}_A, \vec{U}_B, \vec{V}_A, \vec{V}_B$ are orthogonal matrices and $\vec{\Sigma}_A, \vec{\Sigma}_B$ are diagonal matrices whose diagonal entries $\sigma_i(\vec{A})$ and $\sigma_i(\vec{B})$ are non-negative and in descending order. Let $r = \max(\rank(\vec{A}), \,\rank(\vec{B}))$.
    \begin{align*}
    \langle \vec{A}, \vec{B} \rangle &= \trace(\vec{A}'\vec{B})\\
    &= \trace(\vec{V}_A \vec{\Sigma}_A' \vec{U}_A' \vec{U}_B \vec{\Sigma}_B \vec{V}_B')\\
    &= \trace(\vec{V}_B' \vec{V}_A \vec{\Sigma}_A' \vec{U}_A' \vec{U}_B \vec{\Sigma}_B) &&\text{cyclic property of trace}
    \end{align*}
    Note that since $\vec{U}_A, \vec{U}_B, \vec{V}_A, \vec{V}_B$ are all orthogonal matrices, $\Vert \vec{U}_A' \vec{U}_B \Vert_2 \leq \Vert \vec{U}_A' \Vert_2 \Vert \vec{U}_B \Vert_2 = 1$, and in a similar way, $\Vert \vec{V}_B' \vec{V}_A \Vert_2 \leq 1$.
    \begin{align*}
    \langle \vec{A}, \vec{B} \rangle &= \trace(\vec{V}_B' \vec{V}_A \vec{\Sigma}_A' \vec{U}_A' \vec{U}_B \vec{\Sigma}_B)\\
    &= \sum_{i = 1}^{r} \sum_{j = 1}^{r} z_{ij} \cdot \sigma_i(\vec{A}) \sigma_j(\vec{B}) &&\text{where $\sum_{i = 1}^{r} |z_{ij}| \leq 1$, $\sum_{j = 1}^{r} |z_{ij}| \leq 1$}\\
    &\leq \sum_{i = 1}^{r} \sigma_i(\vec{A}) \, \sigma_i(\vec{B}) &&\text{equality $\iff z_{ij}$}
    \begin{cases}
    1, &         \text{if } i=j,\\
    0, &         \text{if } i\neq j.
    \end{cases}\\
    &\leq \left(\sum_{i = 1}^r \sigma^2_i(\vec{A})\right)^{\frac{1}{2}} \left(\sum_{i = 1}^r \sigma^2_i(\vec{B})\right)^{\frac{1}{2}} &&\text{Cauchy-Schwarz Inequality}\\
    &= \Vert \vec{A} \Vert_F \Vert \vec{B} \Vert_F &&\quad\qed 
    \end{align*}
\end{proof}

The equality conditions for the above requires $z_{ii}= 1, \forall i$ as the $\sigma_i$ are in descending order. This implies that $\vec{U}_A' \vec{U}_B$ and $\vec{V}_B' \vec{V}_A$ are identity matrices, which requires $\vec{U}_A = \vec{U}_B$ and $\vec{V}_A = \vec{V}_B$, i.e. $\vec{A}$ and $\vec{B}$ share the same singular vectors. Equality under Cauchy-Schwarz requires the singular values to be scalars of one another: $\sigma_i(\vec{A}) = s \cdot \sigma_i(\vec{B})$ for the same scalar $s > 0, \forall i$.

This proposition is significant as it gives us upper bounds for the inner product and equality conditions to achieve this upper bound. Applying this result to the stacked Jacobian matrix $\overline{\vec{J}}_N$ gives us the following:
    \begin{align*}
    \Vert \overline{\vec{J}}_N \Vert_F^2 &= \trace(\overline{\vec{J}}_N'\overline{\vec{J}}_N)\\
    &= \trace \left(\sum_{i = 1}^N \sum_{j = 1}^N \vec{J}_f (\vec{x}_i)' \vec{J}_f(\vec{x}_j) \right)\\
    &= \sum_{i, j} \trace (\vec{J}_f (\vec{x}_i)', \vec{J}_f(\vec{x}_j))\\
    &= \sum_{i, j} \langle \vec{J}_f (\vec{x}_i), \vec{J}_f(\vec{x}_j) \rangle  &&\text{Frobenius inner product}\\
    &\leq \sum_{i, j} \Vert \vec{J}_f (\vec{x}_i) \Vert_F \Vert \vec{J}_f (\vec{x}_j) \Vert_F  &&\text{Proposition \ref{prop:frob}}
    \end{align*}
With equality if and only if, for all pairs of inputs $(\vec{x}_i, \vec{x}_j)$, we have $\vec{J}_{f} (\vec{x}_i)$ and $\vec{J}_{f} (\vec{x}_j)$ sharing singular vectors and their corresponding singular values are constant up to a fixed scalar $s > 0$.

Our result can be summarized with the following equation:
\begin{equation}
\label{eq:ineq}
\Vert \overline{\vec{J}}_N \Vert_F  \leq \left( \sum_{i, j} \Vert \vec{J}_f (\vec{x}_i) \Vert_F \Vert \vec{J}_f (\vec{x}_j) \Vert_F \right)^{\frac{1}{2}}
\end{equation}
From a defense perspective, this shows that regularizing the Frobenius of the Jacobian for the $\vec{x}_i$ decreases the total Frobenius norm of the stacked Jacobian and hinders the overall effectiveness of a UAP. Thus, data-dependent Jacobian regularization across inputs should make it significantly more difficult to generate effective UAPs.

\subsection{Measuring Alignment of Jacobians}
To measure the alignment between Jacobians of two distinct inputs, we use the \textbf{cosine similarity} between their respective Jacobians under the inner product induced by the Frobenius norm:
\begin{equation}
\label{eq:cos}
\text{sim}(\vec{x}_i, \vec{x}_j) = \frac{\langle \vec{J}_f (\vec{x}_i), \vec{J}_f(\vec{x}_j) \rangle}{\Vert \vec{J}_f (\vec{x}_i) \Vert_F \Vert \vec{J}_f (\vec{x}_j) \Vert_F} \leq 1
\end{equation}
This is precisely the formula given in Proposition 1, with the above ratio equal to one if and only if the singular vectors of their Jacobians are the same. This shows to us that alignment of Jacobians can be evaluated with this similarity measure. Also, combining this with our findings from Eq.~\ref{eq:ineq}, this ratio allows us to  measure how strongly two inputs share adversarial perturbations.

Although the Jacobian is a first-order derivative, we show in later sections that our Jacobian similarity measure correlates with vulnerability to iterative  UAP attacks. Thus, demonstrating that it is an effective measure to determine the ``universality"  of adversarial vulnerability even against iterative adversaries.

Having a similarity measure like this is beneficial as this allows us to easily determine if two inputs are likely to share adversarial perturbations. This is more advantageous than manually generating adversarial perturbations for each pair of inputs as one would have to consider many additional attack parameters when generating adversarial attacks, including the $\varepsilon$ bounds, chosen $\ell_p$-norm, step size, number of attack iterations, and so on.

%% file: sections/4-experiments.tex
\section{Experiments}

\subsection{Experimental Setup}

\textbf{Models \& Datasets.} We consider the benchmark datasets MNIST \cite{lecun1998gradient} and Fashion-MNIST \cite{xiao2017fashion}. These are widely-used image classification datasets, each with 10 classes, whose images are 28 by 28 pixels, and their pixel values range from 0 to 1. For the neural network architecture, we use a modernized version of LeNet-5 \cite{lecun1998gradient} as detailed in \cite{hoffman2019robust} as it is a commonly used benchmark neural network. We refer to this model as LeNet.

\textbf{Jacobian Regularization.} For training with Jacobian regularization (JR), we optimize the following joint loss and use the algorithm as proposed by \cite{hoffman2019robust}:
\begin{equation}
    \mathcal{L}_{\text{joint}}(\vec{\theta}) =  \mathcal{L}_{\text{train}}(\{\vec{x}_i, \vec{y}_i\}_i, \theta) + \frac{\lambda_{\text{JR}}}{2} \left( \frac{1}{B} \sum_i \Vert \vec{J}(\vec{x}_i) \Vert_F^2 \right)
\end{equation}
where $\theta$ represent the parameters of the model, $\mathcal{L}_{\text{train}}$ is the standard cross-entropy training loss, $\{\vec{x}_i, \vec{y}_i\}$ are input-output pairs from the mini-batch, and $B$ is the mini-batch size. This optimization uses a regularization parameter $\lambda_{\text{JR}}$, which lets us adjust the trade-off between regularization and classification loss.

\textbf{UAP Attacks.} We evaluate the robustness of these models to UAPs generated via iterative stochastic gradient descent with 100 iterations and a batch size of 200. Perturbations are applied under $\ell_{\infty}$-norm constraints. The $\varepsilon$ we consider in our attacks for this norm are from 0.1 to 0.3, this perturbation magnitude is equivalent to 10\%-30\% of the maximum total possible change in pixel values.

We generate untargeted and targeted attacks. For targeted UAPs, we generate one UAP for each of 10 classes of each dataset. Clean and UAP evaluations are done on the entire 10,000 sample test sets.

\textbf{Robustness Metrics.} The effectiveness of untargeted attacks are measured using the \emph{Universal Evasion Rate (UER)}, defined as the proportion of inputs that are misclassified. Targeted UAPs for class $c$ are evaluated according to their \emph{Targeted Success Rate (TSR)}, the proportion of inputs classified as class $c$.

\subsection{Jacobian Regularization Mitigates UAPs}
\begin{figure}
\includegraphics[width=0.5\textwidth]{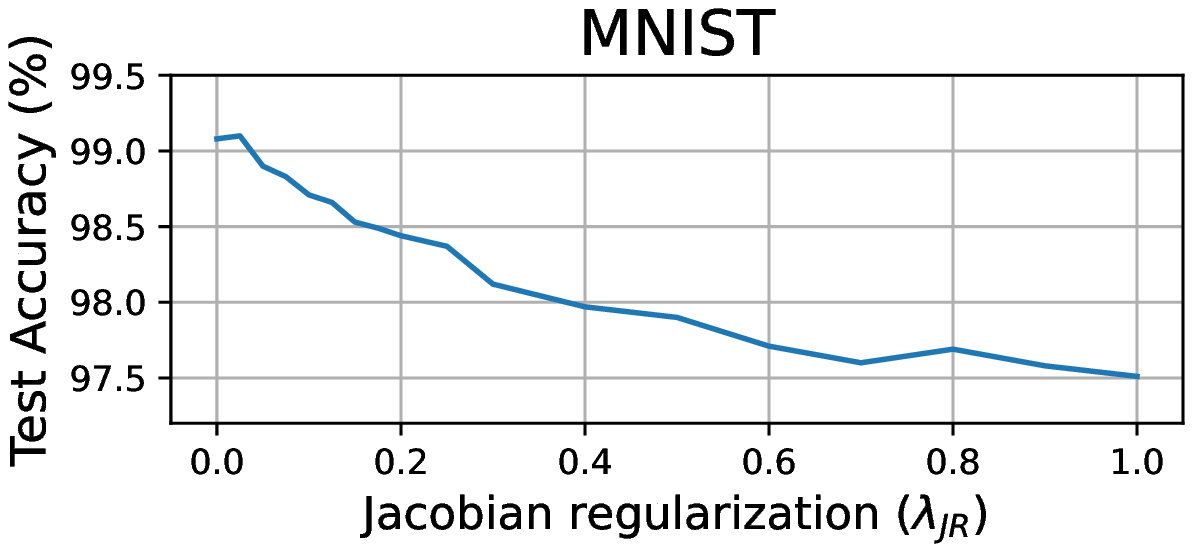}
\includegraphics[width=0.5\textwidth]{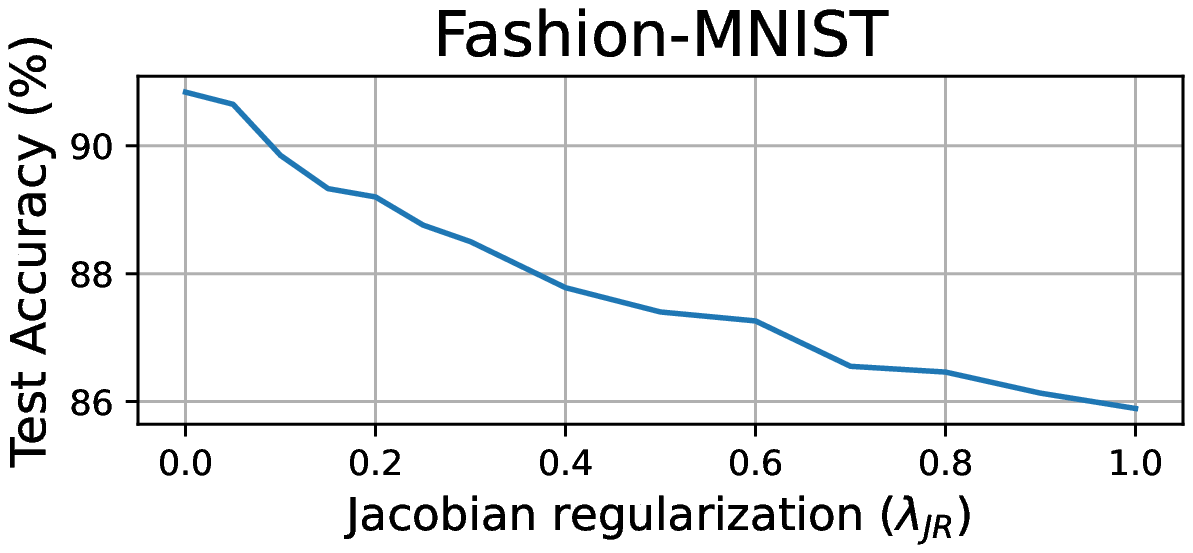}
\caption{Test accuracy of LeNet on MNIST (left) and Fashion-MNIST (right) for various Jacobian regularization strengths $\lambda_{\text{JR}}$.}
\label{fig:testacc}
\end{figure}

Regular training without JR (i.e. $\lambda_{\text{JR}} = 0$) achieves 99.08\% and 90.84\% test accuracy on MNIST and Fashion-MNIST respectively. Fig.~\ref{fig:testacc} shows that increasing the weight of JR decreases the resulting model's test accuracy. Note, however, that this decrease appears to be negligible for very small $\lambda_{\text{JR}} \leq 0.1$.

\begin{figure}
\includegraphics[width=0.5\textwidth]{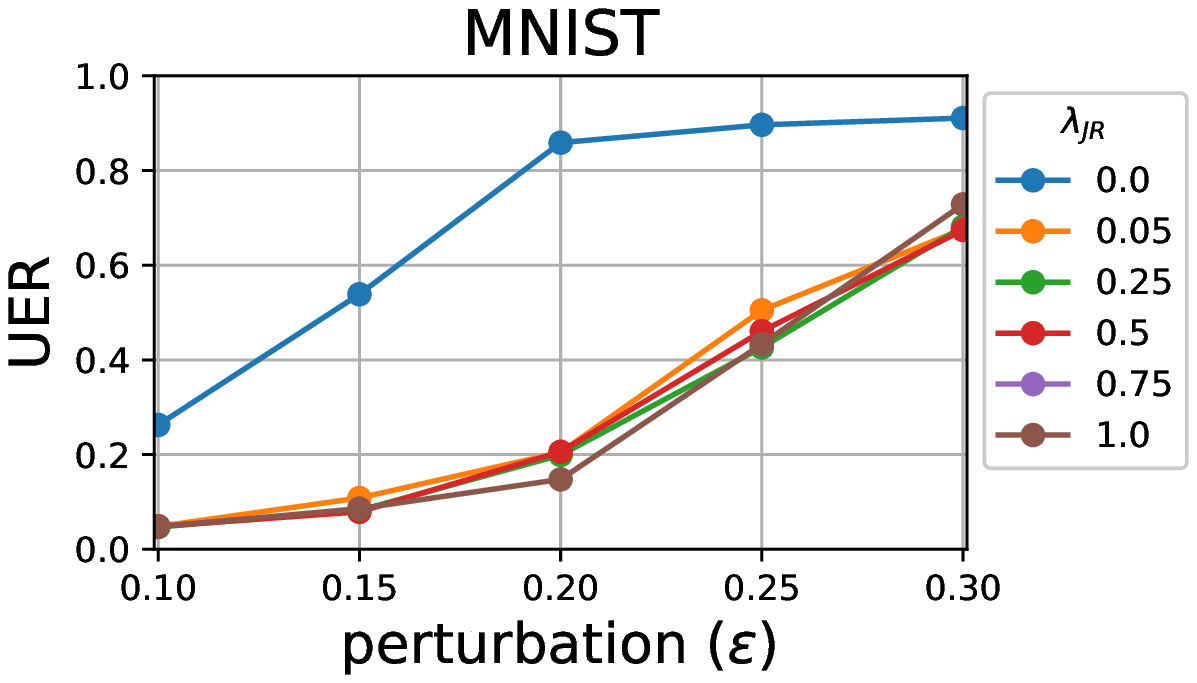}\includegraphics[width=0.5\textwidth]{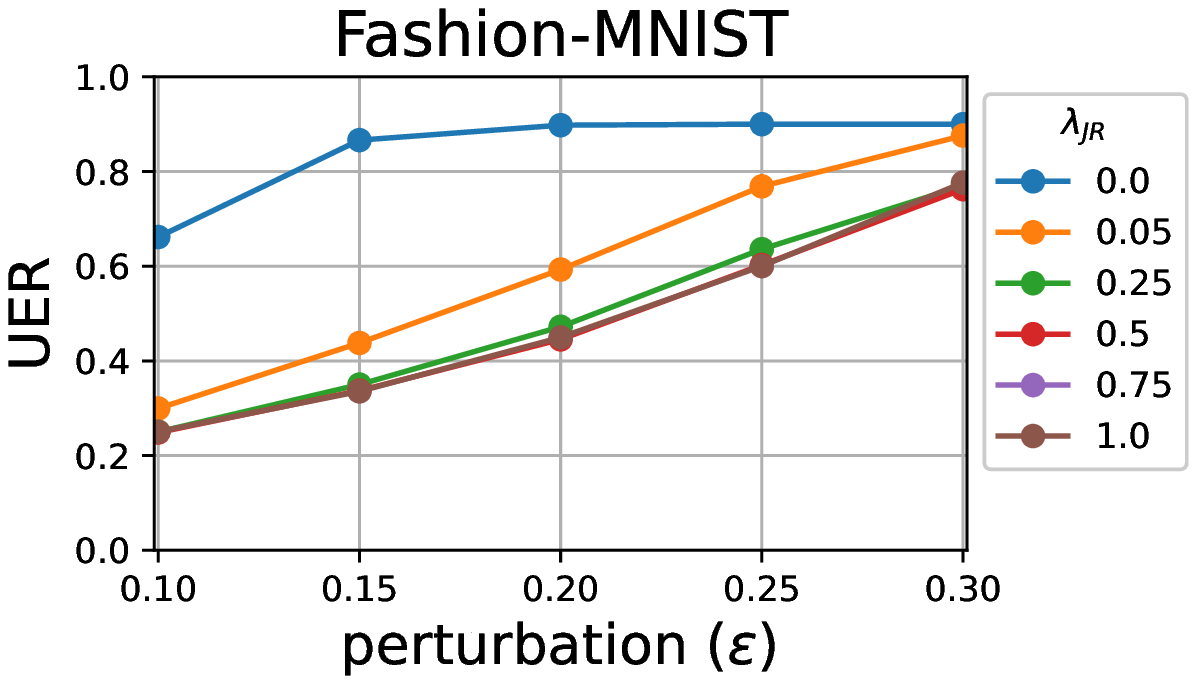}
\caption{Effectiveness of untargeted UAPs for various $\ell_{\infty}$-norm perturbation constraints $\varepsilon$. Plots are shown for various models with different degrees of Jacobian regularization.}
\label{fig:UER}
\end{figure}
\begin{figure}[htb]
\includegraphics[width=0.5\textwidth]{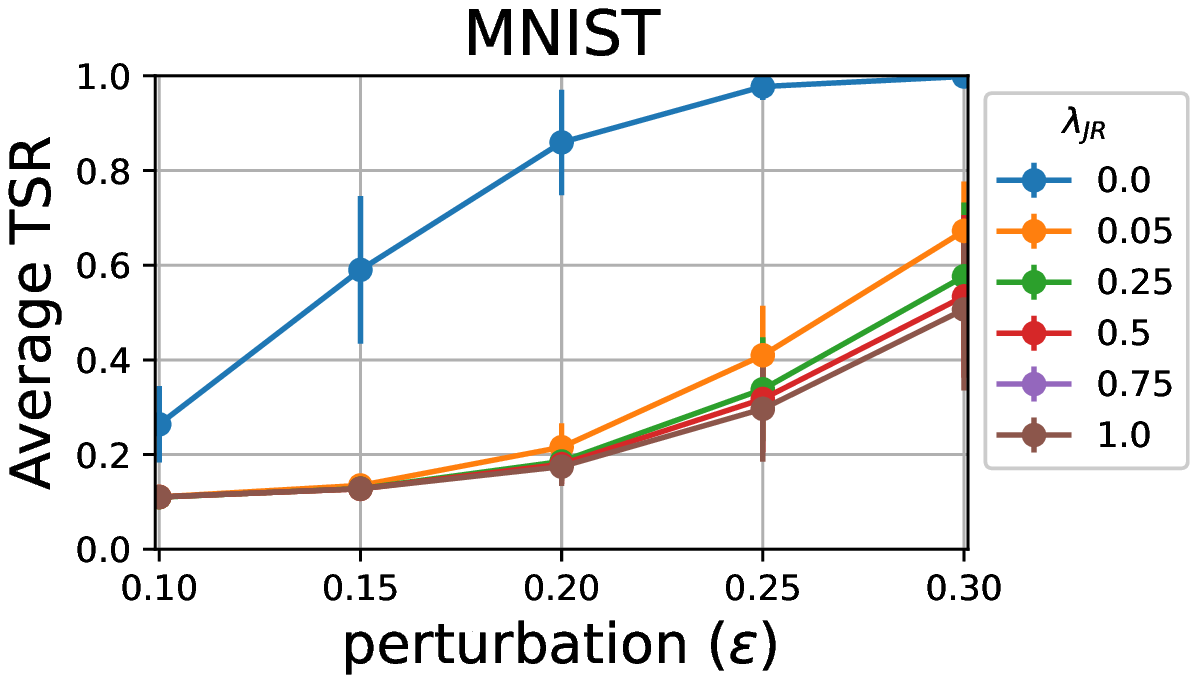} \includegraphics[width=0.5\textwidth]{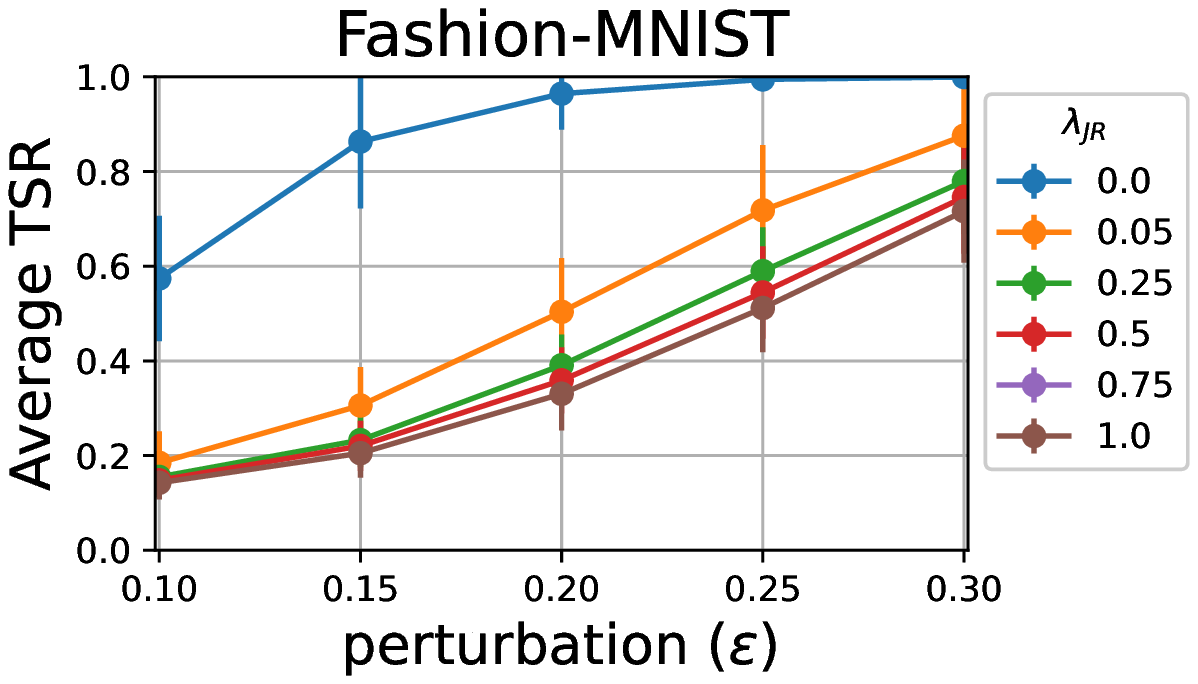}
\caption{Average Targeted Success Rate (TSR) of targeted UAPs generated for each class, with error bars showing standard deviation across UAPs for different classes. Plots are shown for various models with different degrees of Jacobian regularization.}
\label{fig:TSR}
\end{figure}

\textbf{Untargeted UAPs.} Fig.~\ref{fig:UER} presents the effectiveness of our untargeted UAP attacks on different LeNet with varying JR strengths. The regularly trained model is especially vulnerable to UAP attacks on both datasets, with untargeted UAPs achieving above 80\% UER for $\varepsilon \geq 0.2$ on both datasets.

On MNIST, UAP attacks seem to gain reasonable success only after $\varepsilon \geq 0.25$. This is permissible as the adversary perturbs the input by 25\% of its maximum possible value in this case, which entails an enormous change. What is striking is that JR has a protective effect for $\varepsilon \leq 0.2$, even for small amounts of regularization at $\lambda_{\text{JR}} = 0.05$. Here, UAP effectiveness is down from 80\% to 20\% at $\varepsilon = 0.2$. Increasing the strength of the regularization likely has diminishing returns for robustness as stronger regularization also begins to damage clean accuracy, and thus the model's generalization. Fashion-MNIST can be seen to be less robust since it begins with a lower clean accuracy at around 91\%. This means that the model is overall less robust to begin with than the model trained for MNIST, so we can expect it to be less robust to UAP attacks in general. Nonetheless, we still see a protective effect from JR for $\varepsilon \leq 0.15$ even with only a minor degree of regularization $\lambda_{\text{JR}} = 0.05$.

\textbf{Targeted UAPs.} Fig.~\ref{fig:TSR} shows our results for the effectiveness of targeted UAPs. These plots follow a similar trend as with untargeted UAPs, suggesting that JR is able to improve model robustness against a diverse array of UAP attacks and not only against untargeted UAPs.

Even a minor amount of regularization in $\lambda_{\text{JR}} = 0.05$ provides up to a 4-times decrease in effectiveness of UAPs while maintaining the model's performance on the clean test set, as seen in Table~\ref{table:performance}.

\textbf{Comparison with Adversarial Training.} We compare JR with the current state-of-the-art defense against universal attacks: Universal Adversarial Training (UAT) \cite{shafahi2018universal}, where adversarial training is done on UAPs. UAT models in Table~\ref{table:performance} are trained on $\varepsilon = 0.2$ and $\varepsilon = 0.15$ adversaries for MNIST and Fashion-MNIST respectively. Although UAT improves robustness to UAPs compared to standard training, it doubles the test error on both clean datasets. In contrast, JR achieves better robustness than UAT without damaging clean accuracy.

Adversarial training relies on training against specific UAP perturbations. The heuristic quality of UAT makes improving robustness against all possible perturbations computationally difficult. Our results show that regularizing a more general property of the model, in the norm of the Jacobian, leads to better robustness while maintaining accuracy.
\begin{table}[h]
\small
\centering
\caption{Performance metrics (in \%) of LeNet. Jacobian regularization (JR) uses $\lambda_{\text{JR}} = 0.05$. UAP evaluations are for $\ell_{\infty}$-norm attacks at $\varepsilon = 0.2$ for MNIST and $\varepsilon = 0.15$ for Fashion-MNIST. Lowest values indicate the best robustness and are highlighted.}
\begin{tabular}{rccc|ccc}
    & \multicolumn{3}{c}{MNIST} & \multicolumn{3}{|c}{\hspace{1mm} Fashion-MNIST} \\
	 & \hspace{1mm} Standard & \hspace{1mm} UAT \cite{shafahi2018universal} & \hspace{1mm} JR & \hspace{1mm} Standard & \hspace{1mm} UAT  \cite{shafahi2018universal} & \hspace{1mm} JR\\
	\hline
	Test Error & \hspace{1mm} 0.92 & \hspace{1mm} 1.81 & \hspace{1mm} \textbf{0.90} \hspace{1mm} & \hspace{1mm} 9.16 & \hspace{1mm} 16.66 & \hspace{1mm} \textbf{9.15}\\
	Untargeted UER & \hspace{1mm} 85.88 & \hspace{1mm} 27.49 & \hspace{1mm} \textbf{20.47} \hspace{1mm} & \hspace{1mm} 86.63 & \hspace{1mm} 34.10 & \hspace{1mm} \textbf{29.96}\\
	Average TSR & \hspace{1mm} 85.94 & \hspace{1mm} 24.05 & \hspace{1mm} \textbf{21.57} \hspace{1mm} & \hspace{1mm} 86.33 & \hspace{1mm} \textbf{26.64} & \hspace{1mm} 30.59\\
\end{tabular}
\label{table:performance}
\end{table}

\subsection{Jacobian Alignment of Input Pairs}
We now investigate how the cosine similarity of input Jacobians as introduced in Eq.~\ref{eq:cos} correlates with the models' robustness to UAPs. We consider LeNet with Jacobian regularization ($\lambda_{\text{JR}} = 0.05$) and without ($\lambda_{\text{JR}} = 0.0$). The performance of the models on the test sets is the same as the ones in Table~\ref{table:performance}. For each dataset, we take a random subset of 1,000 test set images with a uniform distribution on the output classes. Thus, we measure the similarity for a million input pairs.

\begin{figure}
\includegraphics[width=0.5\textwidth]{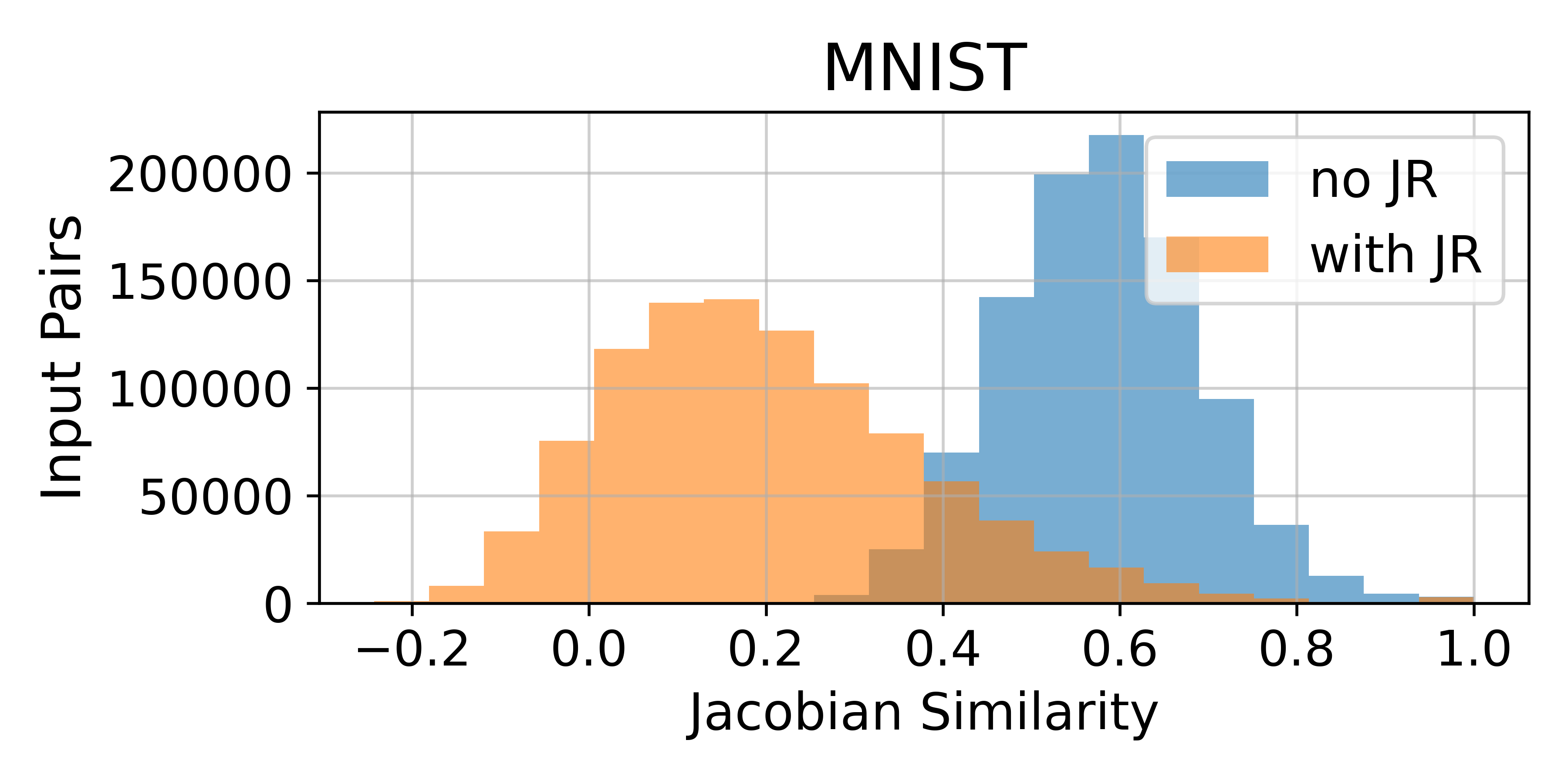}\includegraphics[width=0.5\textwidth]{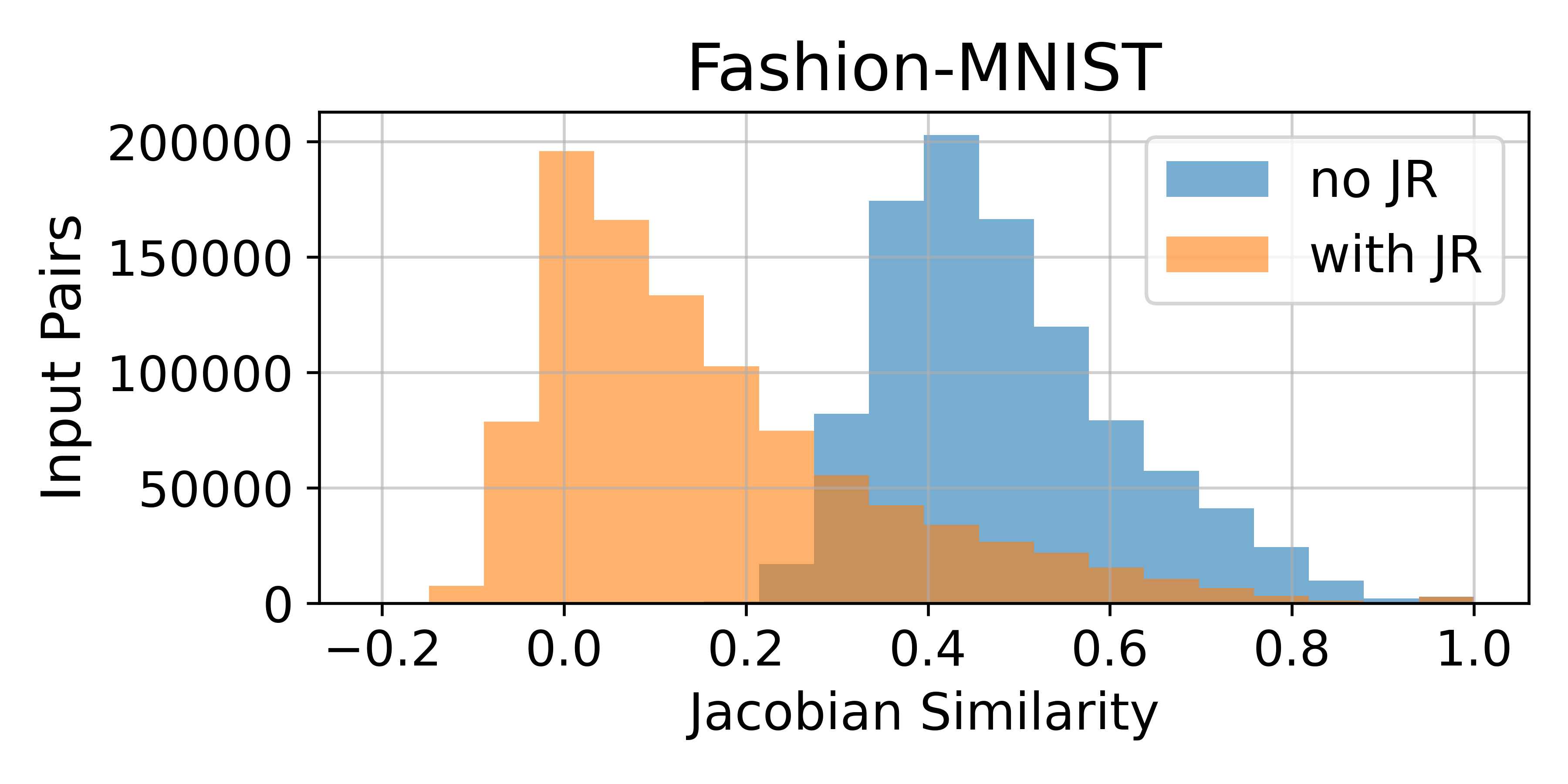}
\caption{Jacobian similarity for pairs of inputs on MNIST (left) and Fashion-MNIST (right) for LeNet with and without Jacobian regularization (JR). Median similarity values on MNIST are 0.18 and 0.58; and on Fashion-MNIST are 0.11 and 0.46 with and without JR respectively.}
\label{fig:jac-sim}
\end{figure}

Fig.~\ref{fig:jac-sim} shows the histogram of the similarity values for the generated random pairs (cosine similarity is bounded in [-1, 1]). We observe that Jacobian regularization significantly reduces the median of the distributions by around 0.35. Although the Jacobian is only a first-order derivative, this greatly correlates with the models' robustness even for iterative stochastic gradient descent UAP attacks. This shows that observing the similarity measure we introduced can help to analyze the strength of shared adversarial perturbations, allowing defenders to better evaluate model robustness against UAPs.

%% file: sections/5-conclusion.tex
\section{Conclusion}
In this work, we are the first to derive upper bounds on the impact of UAPs, we theoretically show and then empirically verify that data-dependent Jacobian regularization significantly reduces the effectiveness of UAPs, and finally we propose cosine similarity of Jacobians to measure the strength of shared adversarial perturbation between inputs.

In contrast to input-specific adversarial examples which have been shown to be difficult to defend against and often incur a notable decline in accuracy to achieve robustness, we show that Jacobian regularization can greatly mitigate the effectiveness of UAPs whilst maintaining clean performance through theoretical bounds and comprehensive empirical results.

These results give us confidence that applying Jacobian regularization to existing models significantly improves robustness to practical and realistic universal attacks at minimal cost to clean accuracy. Additionally, the proposed similarity metric for Jacobians can be used to further diagnose and analyze the vulnerability of models by identifying subsets of inputs with shared adversarial perturbations. Overall, these enable us to put defenses for neural networks against realistic and systemic UAP attacks on a more practical footing.